\documentclass[10pt, a4paper]{article}

\usepackage{lrec-coling2024} 

\usepackage{cite}
\usepackage{amsmath,amssymb,amsfonts}
\usepackage{algorithmic}
\usepackage{graphicx}
\usepackage{textcomp}
\usepackage{xcolor}
\usepackage{CJKutf8}
\usepackage{multirow}
\usepackage{booktabs}

\title{MRC-based Nested Medical NER with Co-prediction and Adaptive Pre-training}

\name{Xiaojing Du$^1$, Hanjie Zhao$^1$\sthanks{~~Xiaojing Du and Hanjie Zhao contributed equally to this research.}~, Danyan Xing$^2$, Yuxiang Jia$^1$\sthanks{~~Yuxiang Jia is the corresponding author.}, Hongying Zan$^1$}

\address{$^1$School of Computer and Artificial Intelligence, Zhengzhou University \\
         $^2$School of Management, Zhengzhou University \\
         \{zzu\_dxj,hjzhao\_zzu,xdy\_0423\}@163.com\\
         \{ieyxjia,iehyzan\}@zzu.edu.cn
         }

\abstract{
In medical information extraction, medical Named Entity Recognition (NER) is indispensable, playing a crucial role in developing medical knowledge graphs, enhancing medical question-answering systems, and analyzing electronic medical records. 
The challenge in medical NER arises from the complex nested structures and sophisticated medical terminologies, distinguishing it from its counterparts in traditional domains. 
In response to these complexities, we propose a medical NER model based on Machine Reading Comprehension (MRC), which uses a task-adaptive pre-training strategy to improve the model's capability in the medical field.
Meanwhile, our model introduces multiple word-pair embeddings and multi-granularity dilated convolution to enhance the model's representation ability and uses a combined predictor of Biaffine and MLP to improve the model's recognition performance.
Experimental evaluations conducted on the CMeEE, a benchmark for Chinese nested medical NER, demonstrate that our proposed model outperforms the compared state-of-the-art (SOTA) models.
 \\ \newline  }

\begin{document}
\begin{CJK*}{UTF8}{gbsn}
\maketitleabstract

\section{Introduction}
With the rapid advancement of medical digitalization, an abundance of medical documentation is being generated, encompassing electronic medical records, medical reports, and various other forms. The extraction of medical information, notably medical named entity recognition (NER), garners increasing significance in applications such as knowledge graph construction, question-answering systems, and automated analysis of electronic medical records. 
Medical NER aims to automatically identify medical entities, including but not limited to body (bod), disease (dis), clinical symptom (sym), medical procedure (pro), medical equipment (equ), drug (dru), and medical examination item (ite), from medical texts.

\begin{figure*}[!ht]  
\centering  
\includegraphics[width=1.0\textwidth]{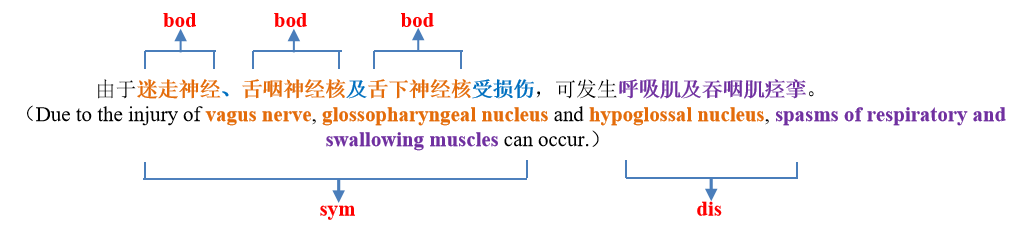}  
\caption{An example of nested entities.}  
\label{fig1}
\end{figure*}

These entities often exhibit lengthy, nested structured, and polysemous, thus presenting considerable challenges to the task of medical NER. For example, as illustrated in Figure \ref{fig1}, the three entities ``迷走神经'' (vagus nerve), ``舌咽神经核'' (glossopharyngeal nucleus) and ``舌下神经核'' (hypoglossal nucleus), denoted as "bod", are nested within the entity ``迷走神经、舌咽神经核及舌下神经核受损伤''(the injury of vagus nerve, glossopharyngeal nucleus and hypoglossal nucleus), denoted as "sym".

To address the challenge of nested NER, we adopt a strategy similar to \citet{li2020unified} and \citet{du2022mrc}, by framing NER as a machine reading comprehension (MRC) task. Like \citet{li2022unified}, we employ an approach that combines the strengths of both Biaffine and Multi-Layer Perceptron (MLP) predictors through joint prediction. Additionally, we introduce a task-adaptive pre-training strategy to fine-tune the original pre-trained model specifically for medical NER.
Our model incorporates several techniques, including Conditional Layer Normalization (CLN), weighted layer fusion, word-pair embeddings, and multi-granularity dilated convolution, all of which have been demonstrated to improve performance. The contributions of this paper can be summarized as follows:
\begin{enumerate}
\item[$\bullet$]We introduce a nested medical NER model based on MRC, featuring the integration of Biaffine and MLP for joint prediction. Additionally, we introduce several enhancements, including multiple word-pair embeddings and multi-granularity dilated convolution, to improve the model's performance.
\item[$\bullet$] We take a task adaptive pre-training strategy to optimize the pre-trained model for medical domain. Entity type embedding is fed into the conditional layer normalization to more effectively utilize entity type information.
\item[$\bullet$]Experimental results on the nested Chinese medical NER corpus CMeEE demonstrate superior performance of our model over existing state-of-the-art (SOTA) models. 
\end{enumerate}

\section{Related Work}

As the medical domain has a high level of complexity and variability among entities, the task of medical NER poses a significant challenge compared to general NER. 
Therefore, employing pre-trained models such as BERT \citep{li2020chinese, qin2021bert} and ELMo \citep{li2020chinese1, wan2020self} has become a common practice to encode input texts in the medical domain. 

To effectively leverage the information in both characters and words, \citet{ji2019hybrid} devise a method that involves constructing a drug dictionary and implementing post-processing rules to modify the entities.
Furthermore, taking into account that radicals, strokes, and glyphs can offer valuable supplementary information alongside words, \citet{zhou2021chinese} propose a BiLSTM-CRF model that operates at the stroke and radical levels to capture the semantic nuances of Chinese characters more comprehensively. 
\citet{yang2022tserl} introduce the TSERL model, which establishes a relationship graph between radicals, characters, and words to enhance NER performance.
Similarly, domain-specific data can be harnessed to augment the performance of medical NER systems. For instance, \citet{liu2021med} develop Med-BERT by pre-training it on a corpus of medical texts, resulting in significant performance enhancements. Additionally, \citet{chen2020improving} propose a model that integrates domain dictionaries and rules with BiLSTM-CRF, showing improved performance in medical NER. This underscores the potential benefits of integrating domain-specific information and rules into existing NER systems.

To address the complexities inherent in nested NER and to incorporate knowledge from entity types, NER has been formulated as an MRC task \citep{li2020unified}. 
Expanding on this, \citet{liu2023fusing} adeptly integrate the interrelations among entity labels utilizing Graph Attention Networks (GATs), thereby fusing label information with textual content to refine NER strategies.
To enhance the exchange of information between the initial and terminal segments of the entity, \citet{cao2021electronic} innovatively apply a Biaffine mechanism to MRC. Further advancements are noted by \citet{zhu2021hitsz}, who synergize sequence labeling and span boundary detection techniques through the implementation of voting strategies. Similarly, \citet{zheng2021chinese} achieve a harmonious ensemble of Conditional Random Fields (CRF) and MRC, showcasing the evolving landscape of NER methodologies.

Multi-task learning represents an alternative approach to enhancing performance. In this framework, the NER model benefits from parameter sharing with models designed for other tasks. \citet{chowdhury2018multitask} explore this approach by considering NER and POS tagging as two concurrent tasks.
Additionally, \citet{du2022mrc} propose an integration of the MRC-CRF model for sequence labeling and the MRC-Biaffine model for span boundary detection within a multi-task learning architecture. Similarly, \citet{luo2020chinese} extend the multi-task learning paradigm to NER across two distinct datasets.

The advent of large language models (LLMs), exemplified by ChatGPT \citep{openai2022chatgpt}, has heralded a new paradigm in entity recognition and an increasing number of studies have been focusing on expansive medical models. 
Notably, ChatDoctor \citep{yunxiang2023chatdoctor} enhances its capabilities through continuous pre-training specifically within the medical domain. Likewise, MedAlpaca \citep{han2023medalpaca} garners positive reviews from experts for its clinical response quality.
In the domain of Chinese medical research, DoctorGLM \citep{xiong2023doctorglm}  leverages data processed by ChatGPT for its training, 
whereas BenTsao \citep{wang2023huatuo} employs fine-tuning with Q\&A data generated by ChatGPT, sourced from CMeKG \citep{CMeKG}. 
Furthermore, HuatuoGPT \citep{zhang2023huatuogpt} improves its performance through a combination of Supervised Fine-Tuning (SFT) and Reinforcement Learning from Human Feedback (RLHF). 
\begin{figure*}[htb]  
\centering  
\includegraphics[width=1.0\textwidth]{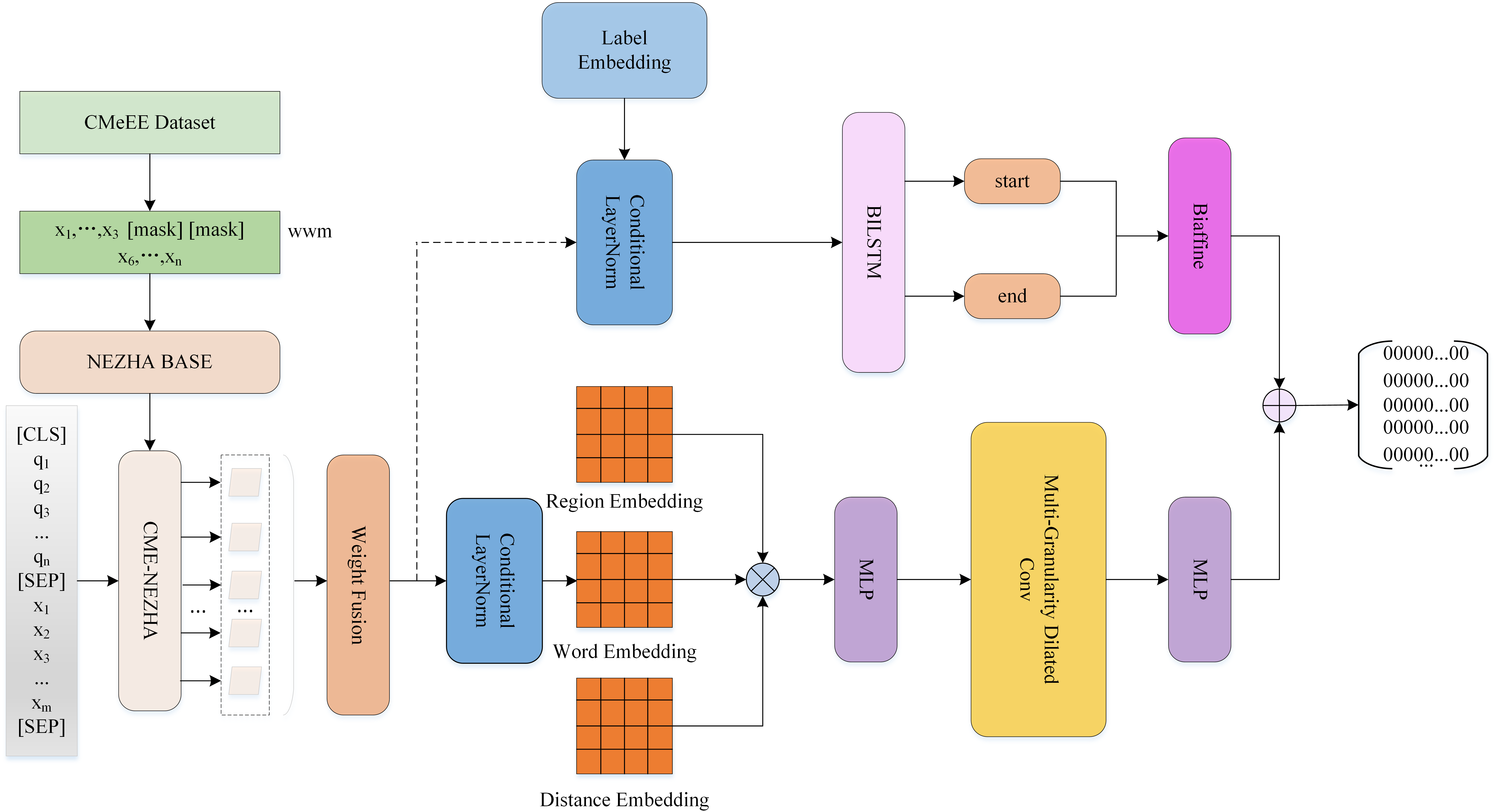}  
\caption{ The architecture of the proposed NER model. The acronym 'wwm' stands for the BERT-Whole Word Masking model.}  
\label{fig2}
\end{figure*}
Additionally, Zhongjing \citep{yang2023zhongjing} enhances its capabilities in Chinese medical consultations by incorporating feedback from medical experts and multi-turn medical dialogues from the real world.

\section{The MRC-CAP Model}
MRC model extracts answer fragments from paragraphs by a given question. Suppose $X$ is the input text, for each entity type $y$, designing a query $q_{y}$, and we can get the triple $(q_{y},y,X)$, which is exactly the $(question,answer,context)$ an MRC model needs. The model only calculates the loss of context during training, and masks the loss of query and padding.

The overall architecture of the proposed MRC-CAP (MRC with Co-prediction and Adaptive Pre-training) model is shown in Figure \ref{fig2}. 
Firstly, incremental training is performed on the existing pre-trained model NEZHA \citep{wei2019nezha} through task adaptive pre-training to obtain the task pre-trained model CME-NEZHA which is more suitable for Chinese medical NER. For each entity type, the input of the model is the concatenation of context and entity description query which we will elaborate on in section \ref{Experiment}. The input is encoded by CME-NEZHA, and then the 12 hidden layers are fused by weights.

Biaffine and MLP are used for joint prediction to enhance the decoding results. For the Biaffine predictor, conditional layer normalization with entity type embedding is used to further leverage entity type knowledge. For the MLP predictor, conditional layer normalization is used to generate word-pair grid representation, combined with other two word-pair embeddings, distance embedding and region embedding, going through a multi-granularity dilated convolution layer for capturing information exchange between distant and close words.

As for the Biaffine prediction branch, the hidden layer sequence after conditional layer normalization goes through a BiLSTM and two nonlinear activation functions to learn the representation of the start and end of the span respectively. Finally, the score of word pair $(x_{i},x_{j})$ is calculated by a Biaffine classifier as follows,

\begin{equation}
	x_{i}=M L P_{\text {start }}\left(h_{i}\right)
\end{equation}

\begin{equation}
	x_{j}=M L P_{\text {end }}\left(h_{j}\right)
\end{equation}

\begin{equation}
	y_{i j}^{\prime}=x_{i}^{{T}} U x_{j}+W\left(x_{i} \oplus x_{j}\right)+b
\end{equation}

where $U$ is a tensor of $N*C*N$, $W$ is a matrix of $2N*C$, $b$ is a bias vector, $N$ is the length of the sentence, and $C$ is the number of entity categories +1 (non-entity).

\begin{table*}[htbp]
\caption{\label{table2} Statistics of entities in CMeEE V1 and V2.}
\begin{center}
\begin{tabular}{l|ccc|ccc}
\toprule
\multirow{2}{*}{Entity} & \multicolumn{3}{c|}{CMeEE V1} & \multicolumn{3}{c}{CMeEE V2} \\ \cline{2-7} 
                               & \#Entity    & Per/\%           & Avg.len            & \#Entity      & Per/\%        & Avg.len            \\ \midrule
bod                            & 23580    & 28.72         & 3.38              & 31467     & 28.94        & 3.36              \\
dis                            & 20778    & 25.31         & 5.34             & 25699      & 23.64     & 5.36              \\
sym                            & 16399    & 19.98         & \textbf{6.70}             & 22415     & 20.62        & \textbf{7.42}              \\
pro                            & 8389     & 10.22         & 5.21              & 13007     & 11.96        & 5.86              \\
dru                            & 5370     & 6.54         & 4.68               & 5945      & 5.47        & 4.78              \\
ite                            & 3504     & 4.27         & 4.29              & 5749       & 5.28      & 4.97              \\
mic                            & 2492     & 3.04        & 4.26             & 2964        & 2.73      & 4.27              \\
equ                            & 1126     & 1.37        & 4.39             & 1053        & 0.96      & 4.73              \\
dep                            & 458      & 0.55        & 2.88              & 431        & 0.40       & 2.55              \\ \midrule
Total                          & 82096    & 100         & 4.89              & 108730      & 100      & 5.17              \\ 
\bottomrule
\end{tabular}
\end{center}
\end{table*}

As described in \citep{li2022unified}, the MLP prediction branch incorporating three word-pair embeddings, including the tensor $V$ of $N*N*d_{h}$
representing word information, the tensor $E^{d}$ of $N*N*d_{E_{d}}$ representing the distance between each pair of words, and the tensor $E^{t}$ of $N*N*d_{E_{t}}$ representing the triangle region information in the word-pair grid. 
The concatenation of three embeddings through an MLP is fed into a multi-granularity dilated convolution layer
with different dilation rate $l$ to capture the interactions between the words of different distances. The computation of one dilated convolution is formulated as:

\begin{equation}
	Q^{l}=\sigma(\operatorname{DConv_{l}}(\operatorname{MLP}\left(\left[{V} ; {E}^{d} ; {E}^{t}\right]\right)))
\end{equation}

where $Q^{l}$ of $N*N*d_{g}$ denotes the output of the dilation
convolution of dilation rate $l$, and $\sigma$ is the GELU \citep{hendrycks2016gaussian} activation function. With the dilation rate 1, 2 and 3, the final word-pair grid representation is $Q=[Q^{1},Q^{2},Q^{3}]$ of ${N*N*3d_{g}}$.
Then an MLP is used to calculate score for word pair $(x_{i},x_{j})$:

\begin{equation}
y_{i j}^{\prime\prime}=M L P\left(Q_{i j}\right)
\end{equation}

With the combination of Biaffine and MLP, we get the co-prediction word pair score $y_{ij}$ :

\begin{equation}
	y_{i j}=\operatorname{Soft} \max \left(y_{i j}^{\prime}+y_{i j}^{\prime\prime}\right)
\end{equation}

In the training stage, we optimize the following cross-entropy loss function:

\begin{equation}
	L_{\text {Biaffine }+M L P}=-\frac{1}{N^{2}} \sum_{i=1}^{N} \sum_{j=1}^{N} \sum_{c=1}^{C} \hat{y}_{i j}^{c} \log y_{i j}^{c}
\end{equation}

\section{Datasets}

The experiment uses the CMeEE V1 and V2 datasets for Chinese nested medical NER. CMeEE V2 is revised from CMeEE V1 by correcting some annotations. 
The texts in CMeEE are extracted from clinical pediatrics textbooks, comprising nine types of entities, including body (bod), disease (dis), clinical symptom (sym), medical procedure (pro), medical equipment (equ), drug (dru), medical examination item (ite), department (dep), and micro-organism (mic).

After analyzing the entity distribution of the two datasets, we observe a similar entity distribution across both datasets with an unbalanced distribution among various types.
"bod", "dis", and "sym" are the dominant types, followed by "pro". 
Table \ref{table2} gives a more detailed comparison of the changes in each entity type and between the two datasets.

Moreover, an analysis is conducted on the proportion of nested versus non-nested entities within the datasets. 
The comparative data in Table \ref{table4} reveals that CMeEE V2 enhances the annotation of nested entities threefold compared to V1, thereby enriching the dataset with more comprehensive nested information. 
\begin{table}[htbp]
\caption{\label{table4} Statistics of nested entities in CMeEE V1 and V2.}
\begin{center}

\begin{tabular}{l|c|c}
\toprule
Entity&CMeEE V1 & CMeEE V2\\ \midrule
\#Flat &73336 & 74160\\ \midrule
\#Nested &8760 & 34570\\ \midrule
Nested/\%&10.67 &31.79\\ \midrule
\#Nested in sym&3808& 14908 \\ \midrule
Nested in sym/\%&23.22& 66.51 \\ 
\bottomrule
\end{tabular}
\end{center}
\end{table}

Take the statement, ``胸部X线透视和胸片可见患侧膈呼吸运动减弱肋膈角变钝。'' (The chest X-ray and chest films reveal diminished diaphragmatic respiratory motion on the affected side and a blunted costophrenic angle.), as an example. 
In V1, the "sym" entity labeled is ``患侧膈呼吸运动减弱'' (diminished diaphragmatic respiratory motion on the affected side). In contrast, V2 annotates the outer "sym" entity ``胸部X线透视和胸片可见患侧膈呼吸运动减弱'' (The chest X-ray and chest films reveal diminished diaphragmatic respiratory motion on the affected side); 
similarly, while V1 identifies ``肋膈角变钝'' (blunted costophrenic angle) as a "sym" entity, V2 enhances this with the inner "bod" entity ``肋膈角'' (costophrenic angle).
Overall, V2 enriches the dataset with a higher number of nested entities through the inclusion of both internally and externally nested entity annotations, thereby offering a more intricate entity structure for the task of medical nested NER.

In examining entities by average length, the entity type "sym" is found to be the longest, with lengths of 6.70 in V1 and 7.42 in V2 respectively. Notably, the nesting rate for "sym" entities, about 2/3, is double that of the dataset overall.
Further examination discloses that "sym" entities with the greatest lengths exhibit intricate structural complexities, typically encompassing entities of other types.

\begin{table}[htbp]
\caption{\label{table5} Statistics of entities nested inside "sym".}
\begin{center}
\begin{tabular}{l|cc|cc}
\toprule
\multirow{2}{*}{Entity} & \multicolumn{2}{c|}{CMeEE V1} & \multicolumn{2}{c}{CMeEE V2}\\ \cline{2-5} 
                               & \#Nested    & Per/\%   
                               & \#Nested     & Per/\%     
                               \\ \midrule
bod                            & 4114    & 85.01        
& 23720     & 78.30                   \\
ite                            & 405     & 8.37         
& 3856       & 12.73                    \\

dis                            & 202    & 4.17        
& 846      & 2.79                 \\
pro                            & 56     & 1.16        
& 1268     &4.19                   \\
dru                            & 27     & 0.56       
& 132      & 0.44                     \\
mic                            & 23     & 0.48         
& 340        & 1.12               \\

equ                            & 12     & 0.25         
& 128        & 0.42                  \\
dep                            & 0      & 0.00         
& 2        & 0.01                 \\ 
\midrule
Total          & 4839    & 100         & 30292        & 100                    \\ \bottomrule
\end{tabular}
\end{center}
\end{table}

Table \ref{table5} presents an analysis of entity types nested within clinical symptom (sym), revealing a widespread occurrence of entity nesting across almost all categories. Notably, the "bod" entities, which are the most prevalent in the dataset, as indicated in Table \ref{table2}, predominate among those nested within "sym".
The distribution of other nested entity types closely mirrors the overall entity distribution observed in Table \ref{table2}. 
Exceptionally, the medical examination item (ite) category demonstrates a significant increase in its proportion of nesting within "sym", securing the second position in the ranking. 
As an illustration, within the entity of type "sym", which states ``小脑延髓池的压力常呈负压'' (The pressure in the fourth ventricle of the cerebellum is usually negative), there is a nested entity of type "ite", specifically referring to ``小脑延髓池的压力'' (The pressure in the fourth ventricle of the cerebellum).

\section{Experiments\label{Experiment}}

\subsection{Experimental Settings}
Below, we provide a detailed description of the experimental setup.

\textbf{Query generation}~
For this study, we incorporate query statements from \citep{du2022mrc} and pair them with examples of relevant entity types such as ``细胞'' (cells), ``皮肤'' (skin) and ``抗体'' (antibodies), which are used as queries to represent the "bod" entity type. 
This is a strategic choice to ensure the accuracy of the query statements in conveying the intended entity types, which in turn provides data with greater informational value for model training.
In application, these query statements are used to enrich the model's understanding of entity types and enhance its overall performance in the NER task. 

\begin{table*}[htbp]
\caption{\label{table6} Query for different entity types in CMeEE \citep{du2022mrc}.}
\begin{center}

\begin{tabular}{l|l}
\toprule
Entity & \multicolumn{1}{c}{Query}                    \\ 
\midrule
bod  & 在文本中找出身体部位，例如细胞、皮肤、抗体     \\       
& Find body parts in the text, for example, cells, skin and antibodies     \\ \midrule
dep        & 在文本中找出科室，例如科、室            \\
& Find departments in the text, for example, department and room           \\ \midrule
dis      &   在文本中找出疾病，例如癌症、病变、炎症、增生、肿瘤 \\
& Find diseases in the text, for example, cancer and pathological changes \\ \midrule
dru       &  在文本中找出药物，例如胶囊、疫苗、剂        \\
& Find drugs in the text, for example, capsule, vaccine and agent        \\ \midrule
equ      &   在文本中找出医疗设备，例如装置、器、导管      \\
& Find medical equipments in the text, for example, device and conduit      \\ \midrule

ite       &  在文本中找出医学检验项目，例如尿常规、血常规    \\
& Find medical examination items in the text, for example, urine routine and \\&blood routine    \\ \midrule
mic      &   在文本中找出微生物，例如病毒、病原体、抗原、核糖  \\
& Find micro-organisms in the text, for example, virus and  pathogen  \\ \midrule
pro      &   在文本中找出医疗程序，例如心电图、病理切片、检测  \\
& Find medical procedure in the text, for example, electrocardiogram and \\&pathological section  \\ \midrule
sym      &   在文本中找出临床表现，例如疼痛、痉挛、异常     \\
& Find clinical manifestations in the text, for example, pain and spasm     \\ \bottomrule
\end{tabular}

\end{center}
\end{table*}

\textbf{Parameter settings}~
To begin with, we perform pre-training on the NEZHA model using the CMeEE dataset, training for 100 epochs to obtain a base model. Then, we fine-tune the base model specifically for NER. 
In the experiments, we set the batch size to 16, the regularization parameter dropout to 0.1, the learning rate for NEZHA to 2e-5, and the other learning rate parameters to 2.5e-3. The maximum text length for the model is set to 200. All experiments are performed on a single NVIDIA RTX3090.

\textbf{Evaluation metrics}~
For evaluation, we employ precise, recall, and F1 scores as performance metrics. In particular, we adopt Micro F1 as a comprehensive metric reflecting the overall recognition performance of the model.
Micro F1, derived from the mean of precision and recall, effectively combines the performance across all categories by weighting each type's sample count.

\subsection{Comparison with Previous Models}

\subsubsection{Baseline Models}
All the baselines we use are as follows:
\textbf{(1)} Lattice-LSTM, Lattice-LSTM+Med-BERT, FLAT-Lattice and Medical-NER are from \citet{liu2021med}. Lattice-LSTM, Lattice-LSTM+Med-BERT and FLAT-Lattice incorporate lexicon to decide entity boundary. Medical NER introduces a big dictionary and pre-trained domain model.
\textbf{(2)} LEAR \citep{yang2021enhanced} independently encodes text and label descriptions and then integrates label knowledge into the text representation through a semantic fusion module. 
\textbf{(3)} MacBERT-large and Human are from \citet{zhang2022cblue}, which is a variant of BERT. Human denotes the annotating result of human.
\textbf{(4)} BERT-CRF, BERT-Biaffine and RICON are from \citet{gu2022delving}. BERT-CRF solves sequence labeling with CRF, BERT-Biaffine detects span boundary with Biaffine, and RICON learns regularity inside entities. 
\textbf{(5)} TsERL \citep{yang2022tserl} constructs a relationship graph between radicals, characters, and words. 
\textbf{(6)} W2NER \citep{li2022unified} proposes a unified word relation classification model for different NER problems. 
\textbf{(7)} MRC-MTL \citep{du2022mrc} integrates MRC-CRF model for sequence labeling and MRC-Biaffine model for span boundary detection into the multi-task learning (MTL) architecture.
\textbf{(8)} FLR-MRC \citep{liu2023fusing} fuses label information with text for NER. 
\textbf{(9)} FFBLEG \citep{cong2023chinese} is based on feature fusion and a bidirectional lattice embedding graph. 
\textbf{(10)} ChatGPT \citep{openai2022chatgpt} and GPT-4 \citep{openai2023gpt4}.
We leverage the API provided by OpenAI \footnote{The results of ChatGPT are obtained during February and March 2024 with official API.}, opting for the GPT-3.5-turbo-16k and GPT-4 model as our baseline.
We conduct a zero-shot experiment and only set the task definition and output format in the prompt template.
Notably, the annotation guidelines specific to the CMeEE dataset are provided.

\subsubsection{Main Results}

\begin{table*}[htbp]
	\caption{\label{table8}Comparison with previous models on CMeEE V1.}
	\begin{center}
		\begin{tabular}{l|ccc}
			\toprule
			Model                           & Pre.  &Rec.  & F1   \\ \midrule
			Lattice-LSTM \citep{liu2021med}       & 57.10          & 43.60          & 49.44          \\
			Lattice-LSTM+Med-BERT \citep{liu2021med} & 56.84          & 47.58          & 51.80          \\ 
			FLAT-Lattice \citep{liu2021med}         & 66.90          & 70.10          & 68.46          \\ 
			Medical NER \citep{liu2021med}            & 66.41          & 70.73          & 68.50          \\ 
			LEAR \citep{yang2021enhanced}           & 65.78          & 65.81          & 65.79          \\ \midrule
			MacBERT-large \citep{zhang2022cblue}          & -              & -              & 62.40          \\ 
			Human \citep{zhang2022cblue}                 & -              & -              & 67.00          \\ 

			BERT-CRF \citep{gu2022delving}             & 58.34          & 64.08          & 61.07          \\ 
			BERT-Biaffine \citep{gu2022delving}       & 64.17          & 61.29          & 62.29          \\
			RICON \citep{gu2022delving}                 & 66.25          & 64.89          & 65.57          \\ 
   			TsERL \citep{yang2022tserl}              & 61.82               & 64.78      & 63.27  \\ 
			W2NER \citep{li2022unified}            & 66.05             & 69.07           & 67.53          \\ 
			MRC-MTL \citep{du2022mrc}              & 66.28              & 70.34          & 68.25          \\ \midrule
   
			FLR-MRC \citep{liu2023fusing}         & 66.79            & 66.25           & 66.52          \\ 
			FFBLEG \citep{cong2023chinese}         & 64.70              & 64.92          & 64.81          \\ \midrule
   		ChatGPT \citep{openai2022chatgpt}        &    42.02        &     32.40     &     36.59    \\ 	
            GPT-4 \citep{openai2023gpt4}        &    39.21      & 50.81        &     44.26   \\ 	

   \midrule
			\textbf{MRC-CAP (Ours)}              & \textbf{67.35}          & \textbf{71.62} & \textbf{69.42} \\ \bottomrule
		\end{tabular}
	\end{center}
\end{table*}

Upon conducting a thorough analysis of our model's performance on the CMeEE V1 dataset in comparison with prior models, we identify substantial advancements in terms of precision, recall, and F1 scores, as highlighted in Table \ref{table8} through bold text.
When compared with the experimental results of large language models, our model still demonstrates certain advantages.

The experimental results for the CMeEE V2 dataset (refer to Table \ref{table10}) are even more remarkable. 
The performance of our model on this dataset surpasses the achievements on V1, with precision increasing from 67.35\% to 77.20\%, and the overall F1 score also achieves significant growth, reaching 77.04\%.
This progress fully reflects the adaptability of our model, especially in dealing with more complex data. 
However, it should be noted that due to the scarcity of research related to CMeEE V2, the comparison results for this part are not listed in Table \ref{table8}.

\begin{table*}[htbp]
	\caption{\label{table10}  Ablation experiments on CMeEE V1 and V2.}
	\begin{center}
		\begin{tabular}{l|ccc|ccc}
			\toprule
			\multirow{2}{*}{Model} & \multicolumn{3}{c|}{CMeEE V1/\%}                                                                 & \multicolumn{3}{c}{CMeEE V2/\%}                                                              \\ \cline{2-7} 
			& \multicolumn{1}{c}{Pre.}      & \multicolumn{1}{c}{Rec.}         & F1        & \multicolumn{1}{c}{Pre.}      & \multicolumn{1}{c}{Rec.}         & F1       \\ \midrule
			MRC-CAP            & \multicolumn{1}{c}{67.35}          & \multicolumn{1}{c}{71.62}          & 69.42          & \multicolumn{1}{c}{77.20}          & \multicolumn{1}{c}{76.88}          & 77.04          \\ 
   
			-AP             & \multicolumn{1}{c}{67.89}          & \multicolumn{1}{c}{69.54}          & 68.70          & \multicolumn{1}{c}{76.97}          & \multicolumn{1}{c}{76.02}          & 76.49          \\ 
			-(AP+MLP)        & \multicolumn{1}{c}{70.71} & \multicolumn{1}{c}{64.09}          & 67.24          & \multicolumn{1}{c}{75.65}          & \multicolumn{1}{c}{74.91}          & 75.28          \\ 
			-(AP+Biaffine)             & \multicolumn{1}{c}{67.64}          & \multicolumn{1}{c}{68.68}          & 68.16          & \multicolumn{1}{c}{75.08}          & \multicolumn{1}{c}{76.42}          & 75.74          \\ 

   -(AP+Biaffine+MLP)                & \multicolumn{1}{c}{67.98}          & \multicolumn{1}{c}{65.87}          & \textbf{66.91}          & \multicolumn{1}{c}{75.39}          & \multicolumn{1}{c}{73.76}          & \textbf{74.56}       

   \\ 
			-(AP+DConv)              & \multicolumn{1}{c}{69.75}          & \multicolumn{1}{c}{65.76}          & 67.69          & \multicolumn{1}{c}{76.79}          & \multicolumn{1}{c}{75.07} & 75.92          \\ 
			-(AP+Region Emb)           & \multicolumn{1}{c}{68.56}          & \multicolumn{1}{c}{67.14} & 67.84 & \multicolumn{1}{c}{76.01} & \multicolumn{1}{c}{76.34}          & 76.17 \\ 

			-(AP+Distance Emb)            & \multicolumn{1}{c}{67.99}          & \multicolumn{1}{c}{67.28} & 67.63 & \multicolumn{1}{c}{76.44} & \multicolumn{1}{c}{75.59}          & 76.01 \\ 
   \bottomrule
		\end{tabular}
	\end{center}
\end{table*}

The experiments conducted affirm the efficacy of our proposed model in enhancing the task of medical nested NER. 
Beyond merely outperforming existing NER models, including ChatGPT, our work validates the effectiveness of the methodologies employed.

\subsection{Ablation Study}
To assess the efficacy of the used modules, our study conducts ablation experiments by omitting the utilized modules within the model across two datasets. 
These experiments demonstrate the effectiveness of our selected modules in enhancing medical NER, with a comprehensive display of improvement in Table \ref{table10}.

Initially, the omission of the adaptive pre-training (AP) module, leds to a decrease in model performance across both datasets. 
We infer that the AP module is instrumental in acquainting the model with an extensive range of medical data, which amplifies its capability in medical NER.
Compared with the single predictors, the joint predictor is observed to enhance the model's recognition ability. 
Notably, the improvement attributed to the MLP is more pronounced compared to Biaffine. 
The exclusion of the joint predictor results in a significant decline in accuracy, 2.51\% on CMeEE V1 and 2.48\% on V2, which we particularly highlight in bold text.

Furthermore, the removal of the multi-granularity dilated convolution (DConv) witnesses a steeper decline on V2, attributed to the dataset's abundance of lengthy and intricate nested entities. 
The experimental outcomes also indicate a decline upon the exclusion of region embeddings or distance embeddings, validating that the integration of embeddings facilitates the learning of more suitable vector representations, thereby elevating the model's performance.


\subsection{Experiments on Various Entity Types\label{Various type}}
To analyze the model's recognition performance on different types of medical entities, we conduct experiments on CMeEE V1. 
Table \ref{table11} illustrates that the entity type "dru" exhibits the highest recognition performance, with an impressive F1 score of 80.50\%. This suggests that most medical drugs are standardized terms with high recognizability. Conversely, the entity type "ite" demonstrates the lowest recognition accuracy at 46.49\%, possibly due to limited data and the majority of "ite" entities being nested within "sym" entities. 

\begin{table}[htbp]
	\caption{\label{table11} Results of different types of NEs on CMeEE V1.}
	\begin{center}
		\begin{tabular}{l|ccc}
			\toprule
			Entity  & Pre. & Rec. & F1  \\ \midrule
			bod   & 67.19        & 65.51     & 66.34       \\ 
			dis   & 79.75        & 77.93     & 78.83       \\ 
			dru   & 76.77        & 84.60     & \textbf{80.50}       \\ 
			dep   & 65.96        & 86.11     & 74.70       \\ 
			equ   & 78.33        & 77.44     & 77.88       \\ 
			ite   & 53.30        & 41.23     & 46.49       \\ 
			mic   & 81.42        & 78.18     & 79.77       \\ 
			pro   & 64.41        & 67.16     & 65.76       \\ 
			sym   & 66.85        & 49.22     & 56.70       \\  \midrule
			Mac-Avg   & 70.44        & 69.71     & 70.07       \\ \bottomrule
		\end{tabular}
	\end{center}
\end{table}

\begin{figure}[htbp]  
	\centering  
	\includegraphics[width=\linewidth]{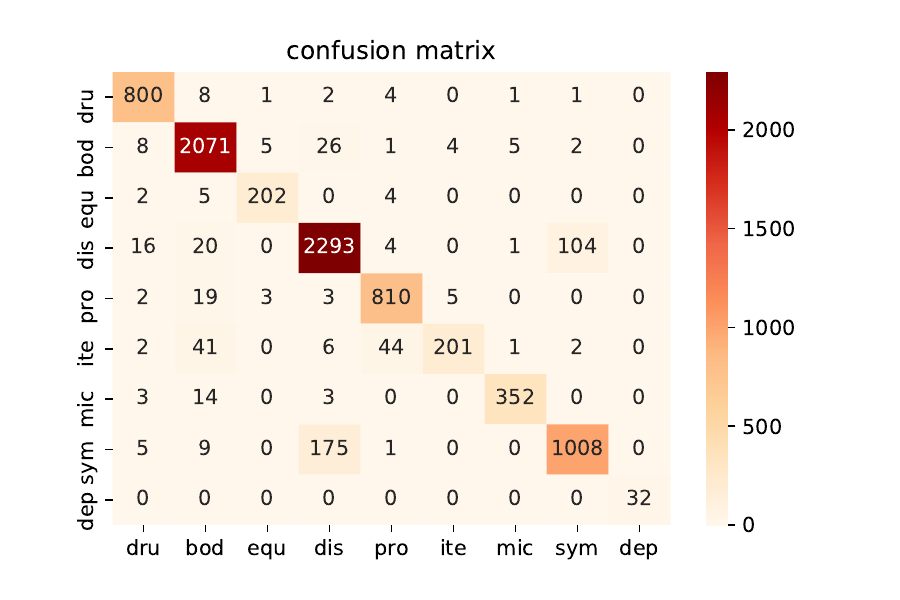}  
	\caption{Confusion Matrix of NER on CMeEE V1.}  
	\label{fig3}
\end{figure}

We devise a confusion matrix to evaluate the model's ability to distinguish between different types of entities. As illustrated in Figure \ref{fig3}, the horizontal axis denotes the correct entity types, referred to as ground truth, while the vertical axis represents the entity types identified by the model. The confusion matrix reveals that the model tends to misclassify "dis" as "sym" most frequently, with 175 out of 215 mistaken instances, and similarly, misclassifying "sym" as "dis" is also a prevalent error (104 out of 109 misclassified "sym" are erroneously labeled as "dis").

A quantitative analysis of the confusion matrix indicates that the model struggles particularly with recognizing complex and lengthy entities, such as "sym" and "dis", which rank among the top two in terms of average length. This difficulty can be attributed, at least in part, to the intricate structures inherent to these entity types.

\subsection{Results of Nested and Flat NER}

\begin{table}[htbp]
	\caption{\label{table12} NER on CMeEE V1(Accuracy/\%). "Inner" and "Outer" denote nested inner entities and nested outer entities, respectively.}
	\begin{center}
		\begin{tabular}{l|c|c}
			\toprule
            Named Entity&MRC&MRC-CAP\\ \midrule
            All & 65.87& 70.43\\
            Flat & 67.54 & 71.75\\
            Nested  & 52.03 & 59.48\\
            Inner & 46.43 & 63.08\\
            Outer & 56.64& 55.10\\
            \bottomrule
		\end{tabular}
	\end{center}
\end{table}

Experimental results for nested entity and flat entity recognition on the CMeEE V1 are presented in Table \ref{table12}. It is evident from the table that the MRC-CAP model achieves higher entity recognition accuracy for both nested inner entities and flat entities compared to traditional MRC-based approaches.
Specifically, the MRC-CAP model demonstrates a 16.65\% and 4.21\% increase in entity recognition accuracy for nested inner entities and flat entities respectively, 
suggesting that enhancing embedded representations enables the model to better capture internal entity information and inter-entity relationships. 
However, the accuracy of nested entity recognition remains considerably lower than that of flat entity recognition, highlighting the need for further advancements in handling nested entity recognition challenges.

\subsection{Case Study}

\begin{table*}[htbp]
\caption{\label{table14}Two cases}
\begin{center}
\begin{tabular}{l|l}
\toprule
Case1         & 结核菌素皮试阳性结核的高危人群，应予以治疗。                                    \\ 
&High risk populations with positive skin test results for tuberculosis should be\\ & treated. \\ \midrule

Golden Entity     & [结核菌素]mic、[皮试]pro、[结核菌素皮试阳性]sym、[结核]dis                                                  \\ \midrule

MRC       & [结核菌素皮试]pro阳性[结核]dis的高危人群，应予以治疗。           \\ \midrule

MRC-CAP & \{[结核菌素]mic[皮试]pro阳性\}sym[结核]dis的高危人群，应予以治疗。 \\ \midrule \midrule

Case2         & 患儿情况好，只1例发生慢性排异及高血压。                                        \\ 
& The condition of the child is good, and only one develops chronic rejection \\ & and hypertension. \\ \midrule
Golden Entity         & [慢性排异]sym、[高血压]sym                                                         \\  \midrule
MRC        & 患儿情况好，只1例发生[慢性排异]dis及[高血压]dis。                                       \\ \midrule
MRC-CAP &  患儿情况好，只1例发生[慢性排异]sym及[高血压]sym。                                       \\ \bottomrule
\end{tabular}
\end{center}
\end{table*}

We select two instances, which are wrongly recognized by MRC but correctly recognized by MRC-CAP. Detailed recognition results are provided in Table \ref{table14}. 
In the first instance, the basic MRC model fails to identify the long entity ``结核菌素皮试阳性'' (tuberculin skin test positive) of type "sym", and erroneously recognizes the two nested entities within it. 
Conversely, MRC-CAP successfully identifies the long "sym" entity and accurately discerns the two inner-nested entities, the "mic" entity ``结核菌素'' (tuberculin) and the "pro" entity ``皮试'' (skin test). 
This suggests that our proposed model has enhanced the ability of MRC to recognize lengthy entities to a certain degree and provides assistance in identifying entities with nested structures.

In the second instance, while the traditional MRC model accurately identifies the boundaries of the entities ``慢性排异'' (chronic rejection) and ``高血压'' (hypertension), it misclassifies "sym" as "dis". Conversely, MRC-CAP correctly identifies the two "sym" entities. Although we acknowledge in section \ref{Various type} that MRC-CAP occasionally misclassifies these two entity types, there is no denying that our proposed model, utilizing a joint predictor, has significantly improved in predicting entity categories compared to traditional MRC models.

\section{Conclusion}

This paper proposes an MRC-based medical NER model for both flat and nested NEs with Biaffine and MLP for joint prediction of NE span, introducing multiple word-pair embeddings and multi-granularity dilated convolution.
To improve domain adaptation of the pre-trained model, we incrementally retrain it with a task-adaptive pre-training strategy.
In addition, entity type embedding, conditional layer normalization, weighted layer fusion and other techniques are employed and show effectiveness. 
Experiments on the nested Chinese medical NER benchmark CMeEE V1 and V2 show that the proposed model outperforms comparative SOTA models. 
In the future, we will incorporate more domain knowledge to improve the performance of the medical NER model and explore potential of LLMs on medical NER task.

\section{Ethics Statement}
There are no ethics-related issues in this paper. We conduct experiments on publicly available datasets. These datasets do not share personal information and do not contain sensitive content that can be harmful to any individual or community.
\section{Acknowledgments}
The authors thank the anonymous reviewers for their insightful comments. This work is mainly supported by the Key Program of the National Natural Science Foundation of China (NSFC) (Grant No.U23A20316), the Key R\&D Project of Hubei Province (Grant No.2021BAA029), and the Major Science and Technology Project of Yunnan Province (Grant No.202102AA100021). The authors are grateful to Zhengzhou Zoneyet Technology Co., Ltd. and Kunming Children's Hospital for their support and cooperation.







\section{Bibliographical References}\label{sec:reference}
\bibliographystyle{lrec-coling2024-natbib}
\bibliography{reference}


\end{CJK*}
\end{document}